%% file: 00_main.tex
\pdfoutput=1
\documentclass[11pt]{article}

\usepackage[]{EACL2023}

\usepackage{times}
\usepackage{latexsym}
\usepackage[T1]{fontenc}
\usepackage[utf8]{inputenc}

\usepackage{microtype}

\usepackage{inconsolata}

\usepackage{amsmath,amsfonts,amssymb}
\usepackage{multirow}
\usepackage{xspace}
\usepackage{subcaption}
\usepackage{esint}
\usepackage{appendix}
\usepackage{etoolbox}
\usepackage{stfloats}
\usepackage{graphicx}
\usepackage[ruled,vlined]{algorithm2e}
\usepackage{enumitem}
\usepackage{algpseudocode}
\usepackage{floatrow}
\usepackage{bm}
\usepackage{bbm}
\newfloatcommand{capbtabbox}{table}[][\FBwidth]

\usepackage{blindtext}
 
\usepackage{wrapfig}
\usepackage{booktabs}
\usepackage{xspace}
\usepackage{placeins}

\usepackage{siunitx}  % for units of measure and data in tables
\newcommand{\system}[1]{\textsc{#1}\xspace}
\newcommand{\data}[1]{\textsc{#1}\xspace}

\newcommand{\geode}{\data{GeoQuery(De)}}
\newcommand{\geoel}{\data{GeoQuery(El)}}
\newcommand{\geoth}{\data{GeoQuery(Th)}}

\newcommand{\geo}{\data{GeoQuery}}
\newcommand{\nlmap}{\data{NLMap}}
\newcommand{\nlmapde}{\data{NLMap(De)}}

\newcommand{\almsp}{\system{AL-MSP}}

\newcommand{\rand}{\system{Random}}

\newcommand{\csse}{\system{CSSE}}

\newcommand{\ssfw}{\system{S2S(FW)}}

\newcommand{\lfsd}{\system{LFSD}}
\newcommand{\lcd}{\system{LCD}}
\newcommand{\lfslcd}{\system{LFS-LC-D}}

\newcommand{\maxcompound}{\system{Max Compound}}

\input{math_comm.tex}

\title{Active Learning for Multilingual Semantic Parser}

\author{Zhuang Li, Gholamreza Haffari\\
 Openstream.AI\\
  {\tt {\{zhuang.li,reza.haffari\}@openstream.com}} 
}

\begin{document}
\maketitle
%\abovedisplayskip=0.5pt
%\abovedisplayshortskip=0.5pt
%\belowdisplayskip=0.5pt
%\belowdisplayshortskip=0.5pt
\begin{abstract}
Current multilingual semantic parsing (MSP) datasets are almost all collected by translating the utterances in the existing datasets from the resource-rich language to the target language. However, manual translation is costly. To reduce the translation effort, this paper proposes the first active learning procedure for MSP (AL-MSP). AL-MSP selects only a subset from the existing datasets to be translated. We also propose a novel selection method that prioritizes the examples diversifying the logical form structures with more lexical choices, and a novel hyperparameter tuning method that needs no extra annotation cost. Our experiments show that AL-MSP significantly reduces translation costs with ideal selection methods. Our selection method with proper hyperparameters yields better parsing performance than the other baselines on two multilingual datasets.
\end{abstract}

\section{Introduction}
\label{sec:intro}
\input{01_intro}
\section{Background}
\label{sec:related}
\input{02_related_work}

\section{Active Learning for MSP}
\input{03_al_method}

\section{Experiments}
\input{04_experiment}
\section{Conclusion}
\label{sec:conclusion}
\input{06_conclusion}
% Entries for the entire Anthology, followed by custom entries
\bibliography{anthology,custom}
\bibliographystyle{acl_natbib}
\clearpage
\newpage
\appendix

\section{Appendix}
\label{sec:appendix}
\input{07_appendix}

\end{document}

%% file: math_comm.tex
\usepackage{amsthm,amsmath,amsfonts,bm,xspace}
\usepackage{color}

\newcommand{\comment}[1]{}

\def\eqref#1{(\ref{#1})}
\def\1{\bm{1}}

\def\vc{{\bm{c}}}

\def\ve{{\bm{e}}}

\def\vx{{\bm{x}}}
\def\vy{{\bm{y}}}

\DeclareMathAlphabet{\mathsfit}{\encodingdefault}{\sfdefault}{m}{sl}
\SetMathAlphabet{\mathsfit}{bold}{\encodingdefault}{\sfdefault}{bx}{n}

%% file: 01_intro.tex
Multilingual semantic parsing converts multilingual natural language utterances into logical forms (LFs) using a single model. However, there is a severe data imbalance among the MSP datasets. Currently, most semantic parsing datasets are in English, while only a limited number of non-English datasets exist.
%To tackle the data imbalance issue, almost all the current works generate MSP datasets by translating utterances in the existing datasets from the resource-rich language (e.g. English) into other languages~\cite{susanto2017neural,duong2017multilingual,li2021mtop}. However, manual translation is slow and laborious. In such cases, active learning is an excellent solution to lower the translation cost. 
To tackle the data imbalance issue, almost all current efforts build MSP datasets by translating utterances in the existing datasets from the resource-rich language (e.g. English) into other languages~\cite{duong2017multilingual,li2021mtop}. However, manual translation is slow and laborious. In such cases, active learning is an excellent solution to lower the translation cost.

Active learning (AL) is a family of methods that collects training data when the annotation budgets are limited~\cite{lewis1994heterogeneous}. Our work proposes the \textit{first} active learning approach for MSP. Compared to translating the full dataset, AL-MSP aims to select only a subset from the existing dataset to be translated, which significantly reduces the translation cost.

We further study which examples AL-MSP should select to optimize multilingual parsing performance. \citet{oren2021finding} demonstrated that a training set with diverse LF structures significantly enhances compositional generalization of the parsers. Furthermore, our experiments show that the examples with LFs aligned with more diversified lexical variants in the training set considerably improve the performance of multilingual parsing during AL. Motivated by both, we propose a novel strategy for selecting the instances which include diversified LF structures with more lexical choices. Our selection method yields better parsing performance than the other baselines. By translating just 32\% of all examples, the parser achieves comparable performance on multilingual \geo and \nlmap as translating full datasets.

%demonstrating that AL-MSP could significantly reduce the translation effort.

Prior works obtain the hyperparameters of the AL methods by either copying configurations from comparable settings or tuning the hyperparameters on the seed evaluation data~\cite{duong2018active}. However, the former method is not suitable as our AL setting is unique, whereas the second method requires extra annotation costs. In this work, we provide a cost-free method for our AL scenario for obtaining optimal hyperparameters.

Our contributions are i) the first active learning procedure for MSP that reduces the translation effort, ii) an approach that selects examples for getting superior parsing performance, and iii) a hyperparameter tuning method for the selection that does not incur any extra annotation costs. %To the best of our knowledge, we are the first to study active learning for MSP. 

%% file: 02_related_work.tex
\noindent\textbf{Multilingual Semantic Parsing.} A multilingual semantic parser is a parametric model $P_{\theta}(\vy|\vx)$ that estimates the probability of the LF $\vy \in \mathcal{Y}$ conditioned on the natural language utterance $\vx \in \mathcal{X}_{l}$ in an arbitrary language from a language set $l \in \mathcal{L}$. The model is trained on the utterance-LF pairs $\{(\vx_i, \vy_i)\}_{i=1}^{N}\in \mathcal{X}_{L}\times \mathcal{Y}$
%\begin{equation}
%    \argmax_{\theta} \prod_{\vx, \vy \in \mathcal{X}_{L}\times \mathcal{Y}} P_{\theta}(\vy|\vx)
%\end{equation}
where $\mathcal{X}_{L} = \bigcup_{l \in L} \mathcal{X}_{l}$ includes multilingual utterances.
%Instead of directly annotating the logical form for the multilingual utterances, most current works collects MSP data by translating utterances from high-resource language into low-resource languages. Such an annotation method is less expensive as it does not require the annotators to have expert knowledge in logical forms.  %where the risk function $R$ is usually the negative log likelihood function, $-\log p_{\theta}(\vy|\vx)$ in semantic parsing research. , where $\mathcal{X} = \bigcup_{l \in L} \mathcal{X}_l$ includes utterances in different languages $L$

\begin{figure}[ht!]
    \centering
    \includegraphics[width=0.77\textwidth]{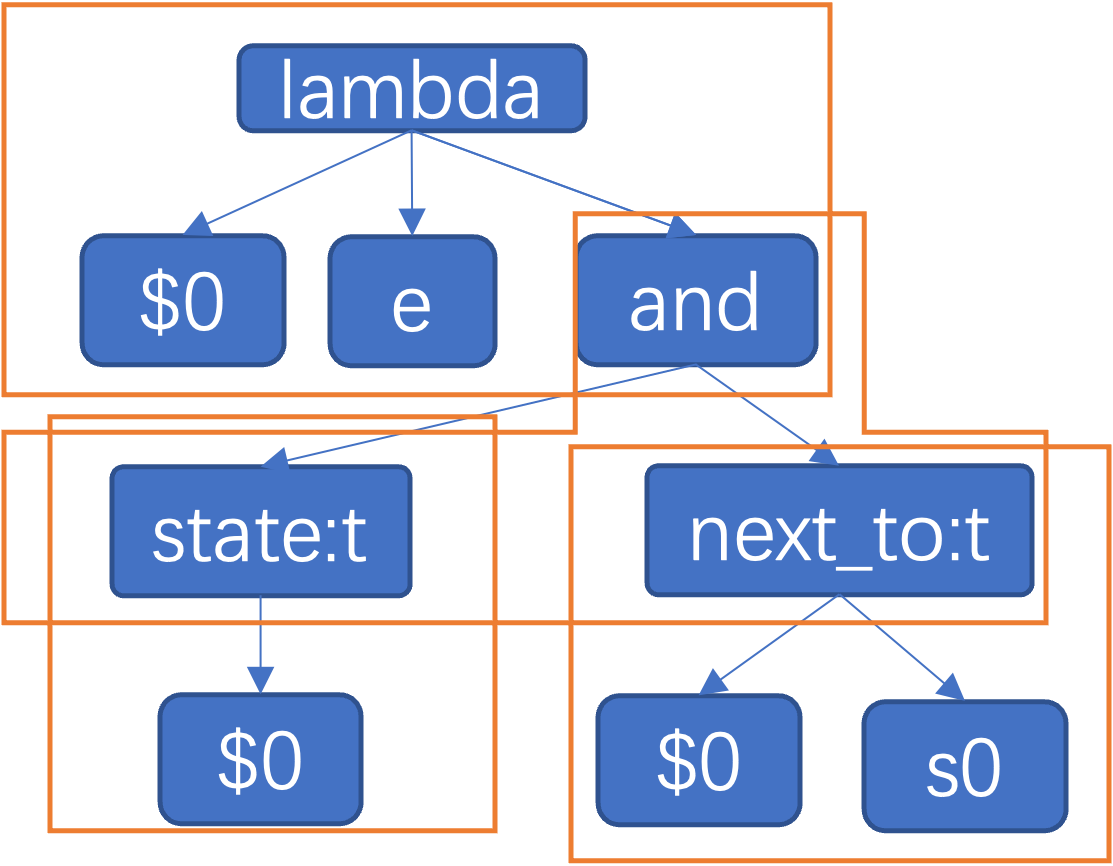}
    \caption{The example of the compounds in an LF tree, \textit{( lambda \$0 e ( and ( state:t \$0 ) ( next\_to:t \$0 s0 ) ) )}.}
    \label{fig:lf_tree_compounds}
\end{figure}

\noindent\textbf{Atoms and Compounds.} Each logical form sequence can be represented as a semantic tree, $\vy = \tau_{\vy}$.~\citet{oren2021finding,shaw2021compositional} define the nodes and sub-trees in $\tau_{\vy}$ as the \textit{atoms} and \textit{compounds}, respectively. Increasing the diversity of the atoms and compounds in the training set improves the parser's compositional generalization~\cite{oren2021finding,li2021total}. For example, an LF ``\textit{( lambda \$0 e ( and ( state:t \$0 ) ( next\_to:t \$0 s0 ) ) )}'' can be expressed as a tree as in Fig.~\ref{fig:lf_tree_compounds}. The atoms are nodes such as ``\textit{lambda}'', ``\textit{\$0}'', ``\textit{e}'' in the LF tree. In this work, the compounds are defined as two-level sub-trees such as ``\textit{( state:t \$0 )}'', ``\textit{( next\_to:t \$0 s0 )}'', ``\textit{( and state:t next\_to:t )}'', and ``\textit{( lambda \$0 e and )}'' in the LF tree.

%~\cite{} both proposed sampling methods to sample the training set with diversied to diversify the LF structures by diversifying the compounds in the sampled examples.

\noindent\textbf{Data Collection for MSP.} 
Prior data collection or active learning works annotates the LFs for the utterances~\cite{duong2018active,sen2020uncertainty} or vice versa~\cite{duong2018active,wang2015building}. But most MSP works~\cite{susanto2017neural,li2021mtop} obtain data by translating existing datasets from high-resource languages into low-resource languages, which is less costly since it does not need annotators' expertise in LFs. Following the same annotation paradigm, our AL does not annotate LFs for multilingual utterances, but instead chooses the utterances to be translated.

%% file: 03_al_method.tex
AL-MSP considers only a bilingual scenario for the proof of concept, while extending our AL method to more than two languages is easy. The goal of AL-MSP is to minimize the human effort in translating utterances while the semantic parser can still achieve a certain level of performance on the bilingual test sets. Starting from a semantic parser initially trained on the dataset $D_s = \{(\vx^{s}_i, \vy_i)\}^{N}_{i=1}$ whose utterances are in the high-resource language $s$, AL-MSP selects $K_q$ examples $\tilde{D}_s = \{(\vx^{s}_i, \vy_i)\}^{K_q}_{i=1}$ from $D_s$, followed by manually translating the utterances in $\tilde{D}_s$ into a target language $t$, denoted by $\tilde{D}_t = t_{s\rightarrow t}(\tilde{D}_s)$, where $\tilde{D}_t = \{(\vx^{t}_i, \vy_i)\}^{K_q}_{i=1}$. The selection criterion is based on our proposed \textit{acquisition} function $\phi(\ve_s)$ scoring each example, $\ve_s = (\vx_s, \vy)$. The parser is re-trained on the union of $\tilde{D}_t$ and $D_s$. There will be $Q$ iterations of selection and re-training until the re-trained parser reaches a good performance on the bilingual test sets $T_{s}$ and $T_{t}$. Algorithm \ref{algo:al} describes our experimental settings in detail.

\input{alg-al-msp}
\subsection{Selection Acquisition}

Our selection strategy selects the untranslated examples which maximize the acquisition scores. The acquisition comprises two individual terms, LF Structure Diversity and Lexical Choice Diversity.

\noindent\textbf{LF Structure Diversity (LFSD).}
We give a simple technique to diversify the LF substructures (atoms and compounds) in the instances. At $q$th iteration, let $D^l_{s}=\bigcup^{q-1}_{i=1} \tilde{D}^{i}_{s}$ denotes all the translated examples and $D^u_{s}=D^{q-1}_{s}$ be the untranslated ones. We partition their union $D^u_{s} \cup D^l_{s}$ into $|D^l_{s}| + K_q$ clusters with Incremental K-means~\cite{cardinal}. Each example $\ve_s = (\vx_s, \vy)$ is featurized by extracting all the atoms and compounds in the LF tree $\tau_{\vy}$, followed by calculating the TF-IDF~\cite{salton1986introduction} value for each atom and compound. Incremental K-means considers each example of $D^l_{s}$ as a fixed clustering centroids and estimates $K_q$ new cluster centroids. For each of the $K_q$ new clusters, we select one example closest to the centroid. 

% The Incremental mechanism guarantees the newly selected examples are not similar to the selected ones.

Such selection strategy is reformulated as selecting $K_q$ examples with the highest acquisition scores one by one at each iteration:
%Our approach ensures that the newly selected examples are not similair with each other and with all selected examples in terms of LF structures . ensures that the newly selected examples are not similair with each other and with all selected examples in terms of LF structures. Such selection strategy is reformulated as selecting $K_q$ examples with the highest acquisition score one by one at each iteration:

\begin{small}
\begin{align}
\label{acq:lcs}
    \phi_{s}(\ve_s) =     
    \begin{cases}
      -||f(\vy) - \vc_{m(\vy)}||^2 & \text{if $m(\vy) \notin \bigcup\limits_{\ve_s \in D^{l}_s} m(\vy)$}\\
      -\infty & \text{Otherwise}
    \end{cases}
\end{align}
\end{small}

\noindent where $f(\cdot)$ is the feature function, $m(\cdot)$ maps each LF into its cluster id and $\vc_i$ is the center embedding of the cluster $i$. As in Algo.~\ref{acq:lcs}, when a new example is chosen, none of its cluster mates will be selected again. The incremental mechanism guarantees the newly selected examples are structurally different from those chosen in previous iterations. Since we use batch-wise AL, we just estimate the clusters once per iteration to save the estimation cost.

\noindent\textbf{Lexical Choice Diversity (LCD).} \lcd aims to select examples whose LFs are aligned with the most diversified lexicons. We achieve this goal by choosing the example maximizing the average entropy of the conditional probability $p(x_s|a)$:

\begin{small}
\begin{align}
     \phi_{c}(\ve_s) &= 
     -\frac{1}{|A_{\vy}|}\sum_{a \in A_{\vy}}\lambda_{a}\sum_{x_s \in V_{s}} p(x_{s}|a) \log p(x_{s}|a) \\
          \lambda_{a} & =     
    \begin{cases}
      1 & \text{if $a \in A_{l}$}\\
      \beta & \text{Otherwise, $0\leq\beta<1$}
    \end{cases}
\end{align}
\end{small}

\noindent {where $a$ is the atom/compound, $A_{\vy}$ is the set of all atoms/compounds extracted from $\vy$, $V_s$ is the vocabulary of the source language,  $A_{l}$ is the set of atoms/compounds in all selected examples until now, and   $p(x_{s}|a)$ is constructed by counting the co-occurrence of $a$ and $x_{s}$ in the source-language training set.
To prevent selecting structurally similar LFs, the score of each selected atom or compound is penalized by a decay weight $\beta$.} %Such decay weight is a hyperparameter betweem 0 and 1. 

{Our intuition has two premises. First, the parser trained on example pairs whose LFs have more lexical choices generalizes better. Second, LFs with more source-language lexical choices will have more target-language lexical choices as well.}

\noindent\textbf{LF Structure and Lexical Choice Diversity (LFS-LC-D).}
We eventually aggregate the two terms to get their joint benefits, $\phi(\vx_s, \vy) = \alpha \phi_s(\vx_s, \vy) + \phi_c(\vx_s, \vy)$, where $\alpha$ is the weight that balances the importance of two terms. We normalize the two terms using quantile normalization~\cite{bolstad2003comparison} in order to conveniently tune $\alpha$.

\noindent\textbf{Hyperparameter Tuning.} Because our setup is unique, we can not copy hyperparameters from existing works. The other efforts~\cite{duong2018active} get hyperparameters by evaluating algorithms on seed annotated data. To tune our AL hyperparameters, $\alpha$ and $\beta$, a straightforward practice using seed data is to sample multiple sets of examples from the source-language data, the target-language counterparts of which are in seed data, by varying different hyperparameter configurations and reveal their translations in the target language, respectively. The parser is trained on different bilingual datasets and evaluated on the \textit{target-side} dev set. We use the one, which results in the best parsing performance, as the experimental configuration.

%Previous efforts get hyperparameters by copying configurations from comparable settings or evaluating algorithms on seed data. However, our work is unique, and seed data annotation costs extra. 
Such a method still requires translation costs on the seed data. We assume if the selected examples help the parser generalize well in parsing source-language utterances, their translations should benefit the parser in parsing target languages. Given this assumption, we propose a \textit{novel} cost-free hyperparameter tuning approach. First, we acquire different sets of source-language samples by varying hyperparameters. Then, we train the parser on each subset and evaluate the parser on the \textit{source-side} dev set. Finally, we use the hyperparameters with the best dev set performance. %Our tuning strategy requires no annotation cost on target language.

%% file: alg-al-msp.tex
\begin{algorithm}[ht]
{\small
\SetKwData{Left}{left}\SetKwData{This}{this}\SetKwData{Up}{up}
\SetKwFunction{Union}{Union}\SetKwFunction{FindCompress}{FindCompress}
\SetKwInOut{Input}{Input}\SetKwInOut{Output}{Output}
\SetAlgoLined
\Input{Initial training set $D^{0} = D^{0}_s$, budget size $K_q$, number of the selection rounds $Q$}
\Output{A well-learned multilingual parser $P_{\theta}(\vy|\vx)$}
Train the parser $P_{\theta}(\vy|\vx)$ on the training set $D^{0}$ \\
\For{$q \gets 1$ to $Q$} 
{
Estimate the acquisition $\phi(\cdot)$ \\
Select a subset $\tilde{D}^{q}_s \in D^{q-1}_s$ of the size $K_q$ based on the acquisition function $\phi(\cdot)$ \\
Translate the utterances in $\tilde{D}^{q}_s$ into the target language, $\tilde{D}^{q}_t = t_{s\rightarrow t}(\tilde{D}^{q}_s)$.\\
Combine the training sets, $D^{q}=D^{q-1} \cup \tilde{D}^{q}_t$\\
Exclude the selected examples $\tilde{D}^{q}_s$ from $D^{q}_s = D^{q-1}_s \setminus \tilde{D}^{q}_s$ \\
Re-train the parser $P_{\theta}(\vy|\vx)$ on $D^{q}$ \\
Evaluate parser performance on test sets $T_s$, $T_t$ 
}
}
\caption{AL-MSP\vspace{-6mm}
}
\label{algo:al}
\end{algorithm}

%% file: 04_experiment.tex
\label{sec:exp}
\noindent\textbf{Datasets.}
We experiment with multilingual \geo and \nlmap. \geo utterances are in English (EN), German (DE), Thai (TH), and Greek (EL); \nlmap utterances are in English and German. Neither corpora include a development set, so we use 20\% of the training sets of \geo and \nlmap in each language as the development sets for tuning the hyperparameters. To simulate AL process, we consider English as the resource-rich language and others as the target languages. After the examples are selected from the English datasets, we reveal their translations in the target languages and add them to the training sets.

%\geo includes 880 utterance-LF pairs about the geography of the U.S. Following conventional splitting as in ~\cite{susanto2017neural}, we use 600 pairs as the training set and 280 pairs as the testing set.~\cite{susanto2017neural} hire human translators to translate the utterances in \geo from English to three target languages, German (De), Thai (Th) and Greek (El). \nlmap\cite{haas2016} includes 1500 training examples and 880 test examples. The English utterances in \nlmap are all translated into German by human translators. 

%To train the multilingual parsers over \geo and \nlmap, we consider English as the \textbf{resource-rich} source language and use Google Translation System to translate the English utterances in \geo into German, Thai, Greek and the ones in \nlmap into German, respectively. The AL methods actively sample English utterances, the human translations of which are obtained from existing human translations in \cite{susanto2017neural} and \cite{haas2016}.
%\minghao{\paragraph{Model Architectures.} run experiments with more model architectures. transformer-base and transformer big should be sufficient.}
\noindent\textbf{AL Setting.} We perform six iterations, accumulatively selecting 1\%,  2\%, 4\%, 8\%, 16\% and 32\% of examples from English \geo and \nlmap.

%Each AL method performs five rounds of query in our AL setting, which accumulatively samples 1\%,  2\%, 4\%, 8\% and 16\% of total examples in \geo and \nlmap. We only performed five rounds as we found the performance of the multilingual parser is saturated after sampling 16\% of examples with most AL methods.
\noindent\textbf{Baselines.} We compare four selection baselines and the oracle setting: i) \textit{Random} picks English utterances randomly to be translated, ii) \textit{S2S (FW)}~\cite{duong2018active} selects examples with the lowest parser confidence on their LFs, iii) \textit{CSSE}~\cite{hu2021phrase} selects the most representative and diversified utterances for machine translation, iv) \textit{Max Compound}~\cite{oren2021finding} selects examples that diversify the atoms and compounds in the LFs, v) \textit{ORACLE} trains the parser on the full bilingual training set.

\noindent\textbf{Evaluation.} We adopt the exact match accuracy of LFs for all the experiments. We only report the parser accuracy on the target languages as we found the influence of new data is negligible to the parser accuracy on English data (See Appendix~\ref{sec:en_res}).

%to estimate the conditional probability $P_{\theta}(\vy|\vx)$.
\noindent\textbf{Base Parser.} We employ BERT-LSTM~\cite{moradshahi2020localizing} as our multilingual parser. Please see Appendix~\ref{app:imple} for its detailed description.
\subsection{Hyperparameter Tuning} 
{Table~\ref{tab:hyper} displays the experiment results with the hyperparameters tuned using only English data (EN) and the hyperparameters tuned using seed data on i) English data plus a small subset (10\% of train data plus development data) in the target language (EN + 10\%), ii) the full bilingual data (EN + full), iii) the same dataset in a different pair of languages from our experiment languages (Diff Lang), iv) a different dataset in the same languages as our experiment (Diff Data). }

%We apply our approach to obtain the best hyperparameters with only English data. For ii) and iii), we select different sets of examples with different hyperparameters from 10\% or full English data and reveal their translations. We evaluate the trained parser on the dev set of the target language. We adopt the same tuning process as in ii) and iii) for iv) and v), except that the target language or the dataset is different. We adopt the hyperparameters which result in best parsing performance on  the dev sets. Table 1 displays the parsing accuracies on the test sets of geo and nlmap in several target languages using our selection technique LFS-LC-D after selecting and translating 16\% of samples with the best hyperparameter configurations obtained from different tuning methods.

\input{tab-hyper}

From Table~\ref{tab:hyper}, we can see our approach takes significantly fewer annotation resources than others to find optimum hyperparameters. Adding more target-language data does not help obtain better hyperparameters, validating our assumption that English data is enough for \lfslcd to obtain good hyperparameters. Surprisingly, the hyperparameters tuned on a different language pair do not significantly worsen the selection choices. However, tuning hyperparameters from other datasets results in inferior parsing performance, which is anticipated as different datasets include different LFs, but the performance of \lfslcd is closely related to the LF structures.
\begin{figure*}[ht!]
\vspace{-6mm}
    \centering
    \includegraphics[width=0.82\textwidth]{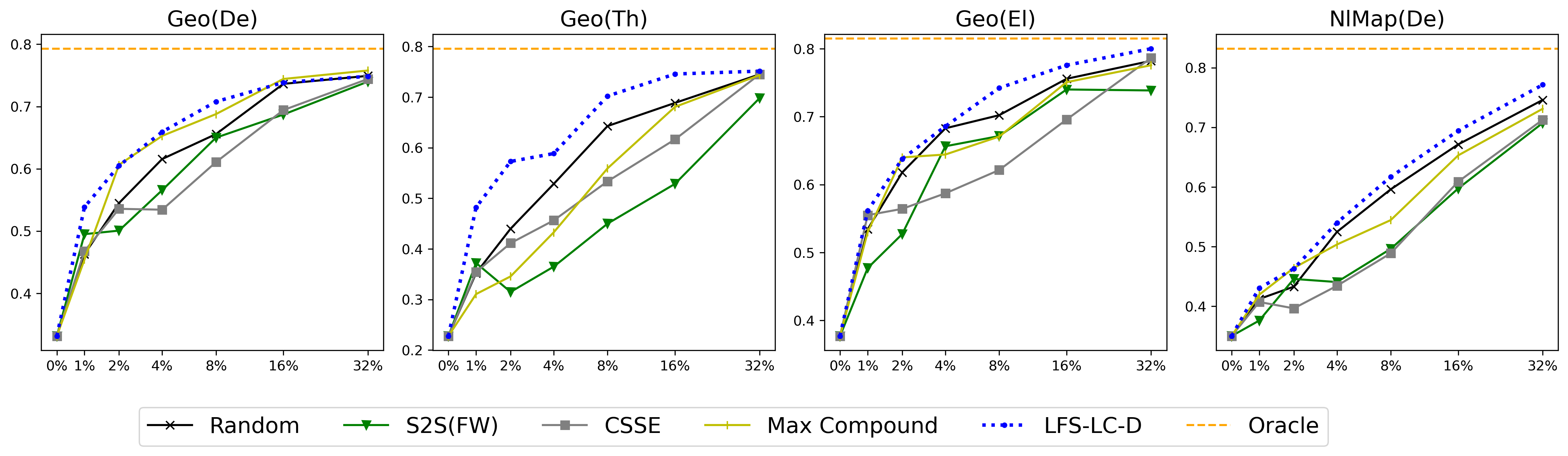}
    \caption{The parsing accuracies at different iterations on the test sets of \geo and \nlmap in German (De), Thai (Th), and Greek (El) using different selection approaches. All experiments are
run five times with different seeds.}
    \label{fig:main}
\end{figure*}

\subsection{Active Learning Results} \noindent\textbf{Effectiveness of AL-MSP.} Fig.~\ref{fig:main} shows that only a small amount of target-language data significantly improves the parsing performance over the zero-shot performance. For example, merely 1\% of training data improves the parsing accuracies by up to 13\%, 12\%, 15\% and 6\% on \geode, \geoth, \geoel and \nlmapde, respectively. With the best selection approach \lfslcd, translating 32\% of instances yields parsing accuracies on multilingual \geo and \nlmap that are comparable to translating the whole dataset, with an accuracy gap of less than 5\%, showing that our \almsp might greatly minimize the translation effort.

\noindent\textbf{Effectiveness of LFS-LC-D.} \lfslcd consistently outperforms alternative baselines on both multilingual datasets when the sampling rate is lower than 32\%. In contrast, \ssfw consistently yields worse parser performance than the other baselines. Our inspection reveals that the parser is confident in instances with similar LFs. \maxcompound diversifies LF structures as \lfslcd, however it does not perform well on \geoth. \csse diversifies utterances yet performs poorly. We hypothesize that diversifying LF structures is more advantageous to the semantic parser than diversifying utterances. \rand also performs consistently across all settings but at a lesser level than \lfslcd.

%, indicating that jointing two terms in \lfslcd might result in a more consistent performance
\begin{figure*}[ht!]
    \centering
    \includegraphics[width=0.82\textwidth]{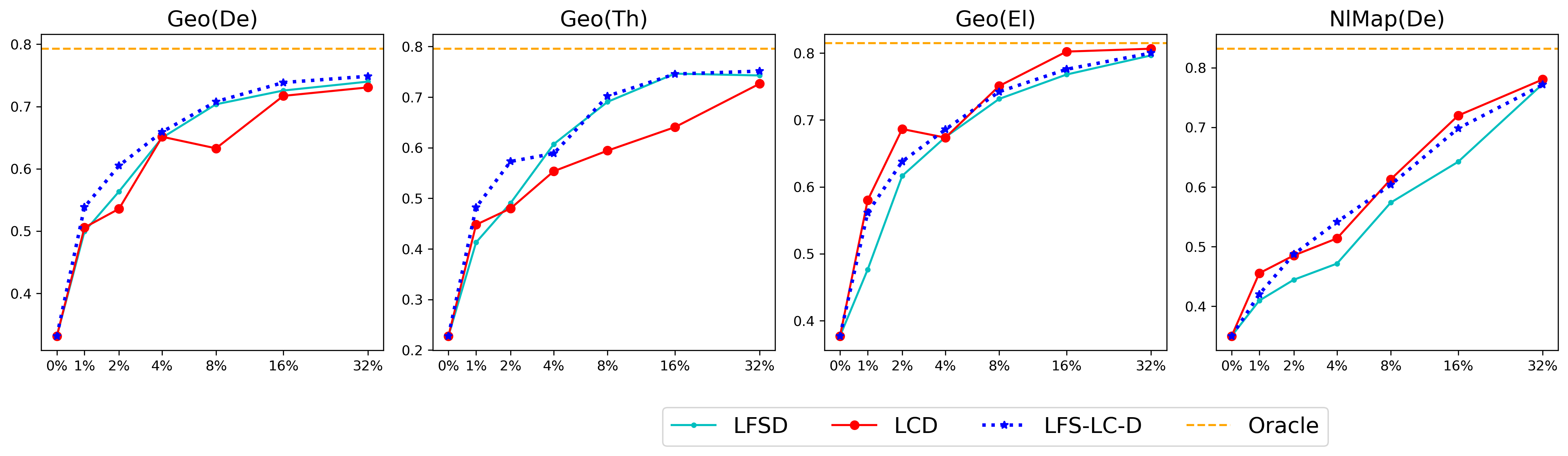}
    \caption{The parsing accuracies at different iterations on the test sets of \geo and \nlmap in German (De), Thai (Th) and Greek (El) using LFSD, LCD, and LFS-LC-D, respectively.}
    \label{fig:abl}
\end{figure*}

\noindent\textbf{Individual Terms of LFS-LC-D.} We also inspect each individual term, \lfsd and \lcd, in \lfslcd. As in Fig.~\ref{fig:abl}, both terms have overall lower performance than \lfslcd, indicating the combination of two terms is necessary. Specifically, \lfsd performs poorly on \nlmap at the low sampling region. We inspect that \nlmap includes 5x more compounds than \geo. Therefore, it is difficult for the small number of chosen examples to encompass all types of compounds. \lcd performs poorly on \geoth. We notice that Thai is an analytic language linguistically distinct from English, German or Greek, so the entropy values of the probability $p(x_s|a)$ over lexicons in Thai (p$=$0.03) is statistically more different to the ones over English than German (p$=$5.80e-30), and Greek (p$=$1.41e-30)\footnote{We use the Student's t-test~\cite{demvsar2006statistical}.}. Overall, the two terms could benefit each other, so \lfslcd performs steadily across different settings. %See Appendix~\ref{sec:abl} for details.

\noindent\textbf{Comparison with Machine Translation.} We also evaluate the parsers that utilize machine translation services. The parsers are trained on a combination of English data and data translated into the target language by Google Translation~\cite{wu2016google}. The accuracy of parsers evaluated on test sets of Geo(De), Geo(Th), Geo(El), and NlMap(De) was 49\%, 58\%, 75\%, and 75\%, respectively. These parsing accuracies are significantly lower than those attained by parsers trained on data provided through human translation, which achieved 80\%, 80\%, 81\%, and 83\%, respectively. This suggests that the performance of the parser is tightly correlated to the quality of the employed machine translation system. Clearly, human translation delivers a greater output quality compared to machine translation. In addition, the results reveal that parsers employing AL methods can easily outperform those employing machine translation methods, particularly when the sampling rate for AL is more than 1\%, 4\%, 8\%, and 32\% in the four data settings.

%% file: tab-hyper.tex
\begin{table}[ht]
\centering
  \resizebox{0.75\textwidth}{!}{%
\begin{tabular}{ccccc}
\toprule
     & \multicolumn{3}{c}{\geo} & \nlmap \\
&DE & TH & EL& DE\\
\midrule
 EN (Ours)  &  73.86 & 74.57 &  77.57 &  69.43\\
 \hline
 EN + 10\%     &  73.86 & 74.57 &  77.57 &  69.02\\
EN + full &  73.86 & 74.57 &  77.14 &  69.43\\
Diff Lang     & 73.86 & 74.04 & 77.57&  -\\
Diff Data    &  71.36 & - &  -&67.72 \\
\bottomrule
\end{tabular}
  }
    \caption{The parsing accuracies on \geo and \nlmap test sets in various target languages after translating 16\% of the English examples selected by \lfslcd with the optimal hyperparameters obtained by different tuning approaches. \vspace{-4mm}}
  \label{tab:hyper}

\end{table}

%% file: 06_conclusion.tex
We conducted the first in-depth empirical study to
investigate active learning for multilingual semantic parsing. In addition, we proposed a method to select examples that maximize MSP performance and a cost-free hyperparameter tuning method. Our experiments showed that our method with the proper hyperparameters selects better examples than the other baselines. Our AL procedure with the ideal example selection significantly reduced the translation effort for the data collection of MSP.
\section*{Limitations}
To reduce annotation costs, existing data collection methods for MSP also utilize machine translation~\cite{moradshahi2020localizing}. Despite the generally lower quality of machine-generated translations compared to human translations, the cost of machine translation services is notably more economical. Our study pioneers the investigation into the feasibility of reducing annotation costs by manually translating only selective portions of the utterance pool. In our work, we provide an initial evaluation of parsers using machine translation versus those using AL methods. Further research is necessary to thoroughly compare these cost-reduction approaches, highlighting their respective advantages and limitations, which we intend to pursue as part of our future work.

%% file: 07_appendix.tex
\subsection{Implementation Details}
\label{app:imple}
\paragraph{BERT-LSTM} BERT-LSTM is a Sequence-to-Sequence model~\cite{sutskever2014sequence} with the XLM-RoBERTa-base~\cite{liu2019roberta} as its encoder and an LSTM~\cite{hochreiter1997long} as its decoder.

\paragraph{Hyperparameters of the Parsers} We tune the hyperparameters of BERT-LSTM on English data. For a fair comparison, we fix the hyperparameters of the parser while evaluating the active learning methods. Specifically, we set the learning rate to 0.001, batch size to 128, LSTM decoder layers to 2, embedding size for the LF token to 256, and epochs to 240 and 120 for the training on \geo and \nlmap, respectively.

\paragraph{Hyperparameters of AL} For tuning the hyperparameters of the active learning method, we grid search the decay weight $\beta$ in 0, 0.25, 0.5, 0.75 and the weight balance rate $\alpha$ in 0.25, 0.5, 0.75, 1. The optimal hyperparameters are 0.75 and 0.75 for all language pairs of \geo and 0.75 and 0.25 for multilingual \nlmap.

In the Diff Lang setting, we assume we can access the data in a language pair other than the experimental one. For selecting English utterances to be translated into German, Thai, and Greek, we tune the hyperparameters on the data of En-Th, En-EL, and En-De pairs, respectively.

In the Diff Data setting, we assume we can access the data in the same language pair as our experimental one but in a different domain with a different type of LF. For selecting English utterances in \geo for translation, we tune the hyperparameters on the bilingual \nlmap. For selecting utterances in \nlmap, we tune the hyperparameters on the \geo in the language pair, En-De.

%\subsection{Ablation of the Individual Terms}
%\label{sec:abl}
%\begin{figure*}[ht!]
%    \centering
%    \includegraphics[width=0.85\textwidth]{figs/geo_abl_res.png}
%    \caption{The parsing accuracies at different iterations on the test sets of \geo and \nlmap in German (De), Thai (Th) and Greek (El) using LFSD, LCD and LFS-LC-D, respectively.}
%    \label{fig:abl}
%\end{figure*}
%As in Fig.~\ref{fig:abl}, the overall performance of each individual term, LFSD or LCD, is lower than the joint selection method, LFS-LC-D, although, in some settings, the individual term can achieve comparable or even higher performance than LFS-LC-D. Therefore, the main benefit of combing two terms is improving the stability of the selection approach so that it can perform steadily across different semantic parsing datasets in different languages.

%We test the statistical significance between the entropy values of conditional probability $p(x_s|a)$ over lexicons in English and the ones in other languages. The p-value calculated with Student's t-test on the language pair of En-Th is 0.03, which is much larger than the p-values on the language pairs of En-De (5.80e-30) and En-El (1.41e-30).
\subsection{Parser Accuracies on English Test Sets}
\label{sec:en_res}
\begin{figure}[ht!]
\vspace{-3mm}
    \centering
    \includegraphics[width=\textwidth]{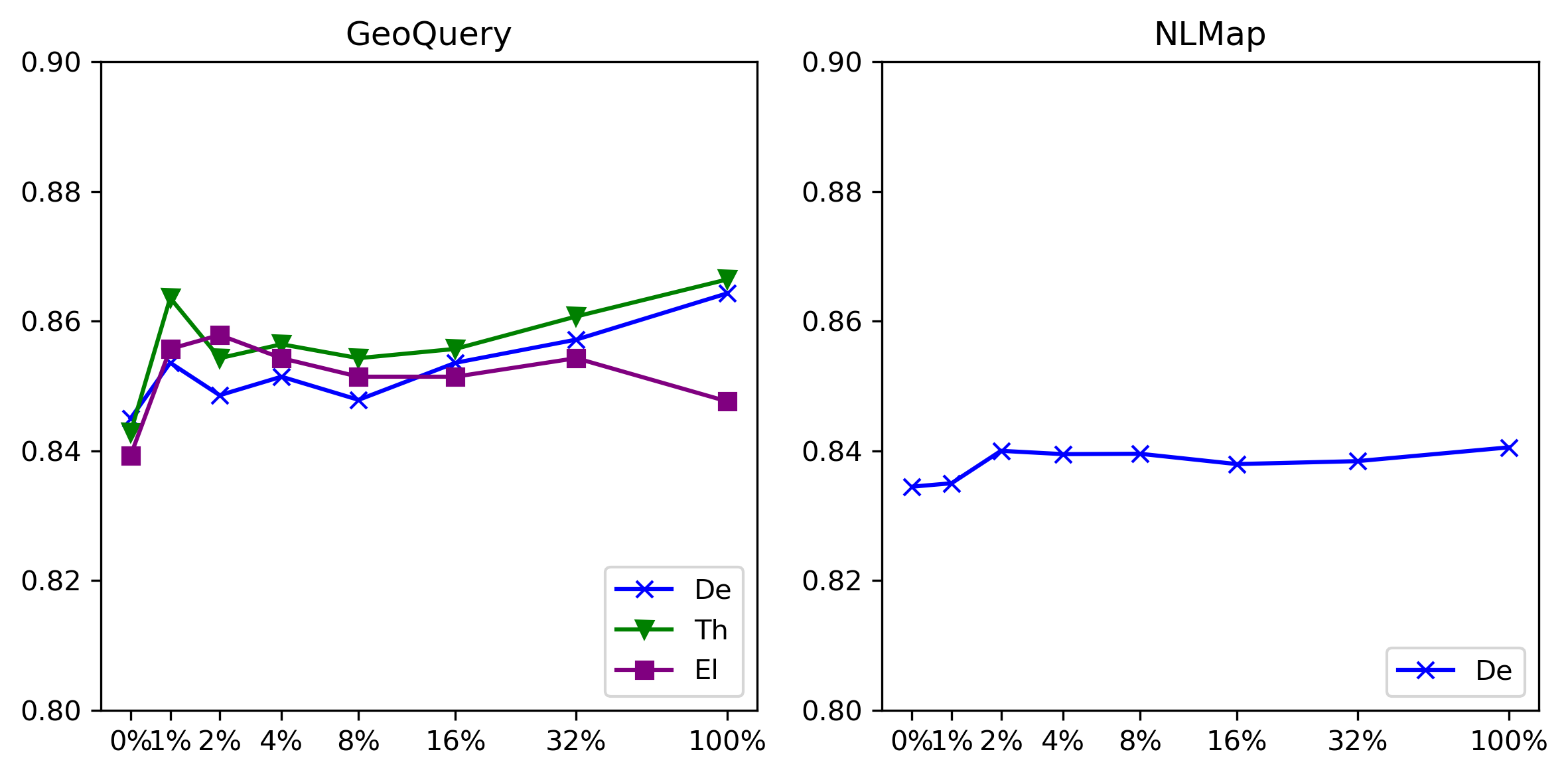}
    \caption{The parsing accuracies at different iterations on the English test sets of \geo and \nlmap after selecting data in German (De), Thai (Th) and Greek (El) using LFS-LC-D, respectively.}
    \label{fig:abl_en_res}
\end{figure}
As in Fig.~\ref{fig:abl_en_res}, training the parser on the data in the target language does not significantly influence the parser's performance on the English test sets. Therefore, in Sec.~\ref{sec:exp}, we only report the experimental results on the test sets in the target languages.